# DBN-Based Combinatorial Resampling for Articulated Object Tracking

**Séverine Dubuisson**   **Christophe Gonzales**   **Xuan Son NGuyen**
Laboratoire d'Informatique de Paris 6 — Université Pierre et Marie Curie
4, place Jussieu, F-75005 Paris
email: `firstname.lastname@upmc.fr`

## Abstract

Particle Filter is an effective solution to track objects in video sequences in complex situations. Its key idea is to estimate the density over the possible states of the object using a weighted sample whose elements are called *particles*. One of its crucial step is a resampling step in which particles are resampled to avoid some degeneracy problem. In this paper, we introduce a new resampling method called *Combinatorial Resampling* that exploits some features of articulated objects to resample over an implicitly created sample of an exponential size better representing the density to estimate. We prove that it is sound and, through experimentations both on challenging synthetic and real video sequences, we show that it outperforms all classical resampling methods both in terms of the quality of its results and in terms of response times.

## 1 INTRODUCTION

Tracking articulated structures with accuracy and within a reasonable time is challenging due to the high complexity of the problem to solve. Actually, the state space of such a problem is inevitably high-dimensional and the estimation of the state of an object thus requires that of many parameters. When the dynamics of the objects is linear or linearizable and when the uncertainties about their position are Gaussian or mixtures of Gaussians, tracking can be performed analytically by Kalman-like Filters [Chen, 2003]. Unfortunately, in practice, such properties seldom hold and people often resort to sampling to approximate solutions of the tracking problem. The Particle Filter (PF) methodology [Gordon et al., 1993] is popular among these approaches and, in this paper, we focus on PF.

PF consists of estimating the density over the states of the tracked object using weighted samples whose elements are possible realizations of the object state and are called *particles*. PF and its variants, e.g., Partition Sampling (PS) [MacCormick and Blake, 1999], all use a resampling step to avoid a problem of degeneracy of the particles, i.e., the case when all but one of the particle's weights are close to zero [Douc et al., 2005]. Without this step, this problem would necessarily occur [Doucet et al., 2001].

A few resampling algorithms are classically used, e.g., Multinomial Resampling [Gordon et al., 1993], Residual Resampling [Liu and Chen, 1998], Stratified and Systematic Resampling [Kitagawa, 1996]. However, these methods have not been designed specifically to deal with articulated objects and, as such, they do not exploit their features. In this paper, we introduce *Combinatorial Resampling*, an algorithm that exploits them to produce better samples by resampling over an implicitly created sample of an exponential size. More precisely, in articulated object tracking, a particle may be thought of as a tuple of the realizations of each "part" of the object and it is often the case that swapping the realizations of a given part among several particles has no impact on the estimated distribution. For instance, in a human body tracking, it may be the case that swapping the positions of the left arm estimated by two particles does not alter the estimated distribution. Given a particle set, Combinatorial Resampling produces implicitly a new set of particles resulting from all such swappings and resamples from it. As such, this new set is of exponential size and acts as a much better description of the state space.

The paper is organized as follows. The next section recalls the basics of articulated object tracking and, in particular, Partitioned Sampling. It also recalls the fundamentals of dynamic Bayesian networks, as our resampling method relies on them. Section 3 presents a short overview of the aforementioned classical resampling approaches and the next one details our new resampling approach and its correctness. Section 5 shows some experimental results both on challenging synthetic and real video sequences. Those highlight the efficiency of our method both in terms of the quality of its results and in terms of response times. Finally, we give some concluding remarks and perspectives.

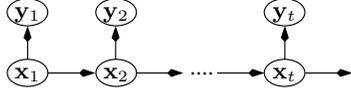

Figure 1: A Markov chain for object tracking.

## 2 ARTICULATED OBJECT TRACKING

In this paper, articulated object tracking consists of estimating a state sequence $\{\mathbf{x}_t\}_{t=1,...,T}$, whose evolution is given by equation $\mathbf{x}_t = \mathbf{f}_t(\mathbf{x}_{t-1}, \mathbf{n}_t^{\mathbf{x}})$, from observations $\{\mathbf{y}_t\}_{t=1,...,T}$ related to the states by $\mathbf{y}_t = \mathbf{h}_t(\mathbf{x}_t, \mathbf{n}_t^{\mathbf{y}})$. Usually, $\mathbf{f}_t$ and $\mathbf{h}_t$ are nonlinear functions, and $\mathbf{n}_t^{\mathbf{x}}$ and $\mathbf{n}_t^{\mathbf{y}}$ are i.i.d. noise sequences. From a probabilistic viewpoint, this problem can be represented by the Markov chain of Fig. 1 and it amounts to estimate, for any $t$, $p(\mathbf{x}_{1:t}|\mathbf{y}_{1:t})$ where $\mathbf{x}_{1:t}$ denotes the tuple $(\mathbf{x}_1, \ldots, \mathbf{x}_t)$. This can be computed iteratively using Eq. (1) and (2), which are referred to as a *prediction step* and a *correction step* respectively.

$$p(\mathbf{x}_{1:t}|\mathbf{y}_{1:t-1}) = p(\mathbf{x}_t|\mathbf{x}_{t-1})p(\mathbf{x}_{1:t-1}|\mathbf{y}_{1:t-1}) \quad (1)$$
$$p(\mathbf{x}_{1:t}|\mathbf{y}_{1:t}) \propto p(\mathbf{y}_t|\mathbf{x}_t)p(\mathbf{x}_{1:t}|\mathbf{y}_{1:t-1}) \quad (2)$$

with $p(\mathbf{x}_t|\mathbf{x}_{t-1})$ the transition corresponding to $\mathbf{f}_t$ and $p(\mathbf{y}_t|\mathbf{x}_t)$ the likelihood corresponding to $\mathbf{h}_t$.

The PF framework [Gordon et al., 1993] approximates the above densities using weighted samples $\{\mathbf{x}_t^{(i)}, w_t^{(i)}\}$, $i = 1, \ldots, N$, where each $\mathbf{x}_t^{(i)}$ is a possible realization of state $\mathbf{x}_t$ called a *particle*. In its prediction step (Eq. (1)), PF propagates the particle set $\{\mathbf{x}_{t-1}^{(i)}, w_{t-1}^{(i)}\}$ using a proposal function $q(\mathbf{x}_t|\mathbf{x}_{1:t-1}^{(i)}, \mathbf{y}_t)$ which may differ from $p(\mathbf{x}_t|\mathbf{x}_{t-1}^{(i)})$ (but, for simplicity, we will assume they do not); in its correction step (2), PF weights the particles using a likelihood function, so that $w_t^{(i)} \propto w_{t-1}^{(i)} p(\mathbf{y}_t|\mathbf{x}_t^{(i)}) \frac{p(\mathbf{x}_t^{(i)}|\mathbf{x}_{t-1}^{(i)})}{q(\mathbf{x}_t^{(i)}|\mathbf{x}_{1:t-1}^{(i)}, \mathbf{y}_t)}$, with $\sum_{i=1}^N w_t^{(i)} = 1$. The particles can then be resampled: those with the highest weights are duplicated while the others are eliminated. The estimation of the posterior density $p(\mathbf{x}_t|\mathbf{y}_{1:t})$ is then given by $\sum_{i=1}^N w_t^{(i)} \delta_{\mathbf{x}_t^{(i)}}(\mathbf{x}_t)$, where $\delta_{\mathbf{x}_t^{(i)}}$ are Dirac masses centered on particles $\mathbf{x}_t^{(i)}$.

As shown in [MacCormick and Isard, 2000], the number of particles necessary for a good estimation of the above densities grows exponentially with the dimension of the state space, hence making PF's basic scheme unusable in real-time for articulated object tracking. To cope with this problem, different variants of PF have been proposed, ranging from local search-based methods like the *Annealed Particle Filter* [Deutscher and Reid, 2005, Gall, 2005] and hierarchical-refining methods [Chang and Lin, 2010] to decomposition techniques like *Partitioned Sampling* (PS) [MacCormick and Blake, 1999] and its siblings [Rose et al., 2008, Besada-Portas et al., 2009]. Here, we focus on decomposition-based particle filters like PS.

PS's key idea is that the state and observation spaces $\mathcal{X}$ and $\mathcal{Y}$ can often be naturally decomposed as $\mathcal{X} = \mathcal{X}^1 \times \cdots \times \mathcal{X}^P$ and $\mathcal{Y} = \mathcal{Y}^1 \times \cdots \times \mathcal{Y}^P$ where each $\mathcal{X}^j$ represents some "part" of the object. For instance, on Fig. 2, a human body is decomposed as 6 parts (head, torso, etc.) numbered from 1 to 6. The state of the $j$th part at time $t$ is denoted $\mathbf{x}_t^j$. Then, by exploiting conditional independences among different subspaces $(\mathcal{X}^j, \mathcal{Y}^j)$, PS estimates $p(\mathbf{x}_{1:t}|\mathbf{y}_{1:t})$ using only sequential applications of PF over $(\mathcal{X}^j, \mathcal{Y}^j)$. For instance, on Fig. 2, given the position of the torso, the left and right arm positions may be independent so, after applying PF on the torso, PS can apply it sequentially to the left and right arms and still compute a correct estimation of $p(\mathbf{x}_{1:t}|\mathbf{y}_{1:t})$. As the $(\mathcal{X}^j, \mathcal{Y}^j)$ subspaces are "smaller" than $(\mathcal{X}, \mathcal{Y})$, the distributions to estimate at each iteration of PF have fewer parameters than those defined on $(\mathcal{X}, \mathcal{Y})$, which significantly reduces the number of particles needed for their estimation and, thus, speeds up the computations.

The exploitation of the conditional independences among the $(\mathcal{X}^j, \mathcal{Y}^j)$ leads to generalizing the Markov chain of Fig. 1 by the Dynamic Bayesian Network (DBN) of Fig. 2

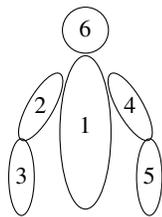
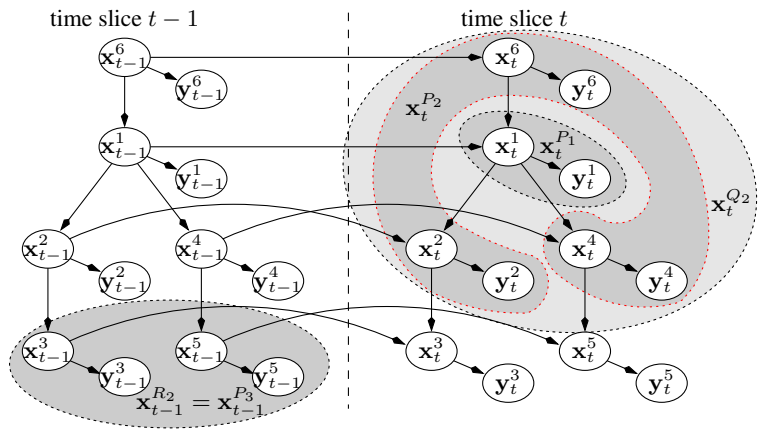

A human body: part 1 corresponds to the torso, parts 2 and 3 to the left arm, parts 4 and 5 to the right arm and part 6 to the head. On the right side of the figure, the corresponding DBN: to the $j$th part corresponds a pair of state and observation variables $\mathbf{x}_t^j, \mathbf{y}_t^j$. The arcs show the dependences between variables, including between different time slices.

Figure 2: A Dynamic Bayesian Network.

[Murphy, 2002], in which the global state $\mathbf{x}_t$ of the object is more finely described as the set of states $\mathbf{x}_t^j$ of each part of the object. The semantics of DBNs is similar to that of Markov chains: the arcs correspond to probabilistic dependences and the joint distribution over all the nodes in the network is equal to the product of the distributions of each node conditionally to its parents in the graph.

The resampling scheme we introduce in this paper, i.e., Combinatorial Resampling, is designed to be part of PS-like algorithms and relies on DBNs. Therefore, we shall now formalize PS in terms of operations over DBNs. For this purpose, for any set $J = \{j_1, \ldots, j_k\}$, let $\mathbf{x}_t^J$ denote the tuple $(\mathbf{x}_t^{j_1}, \ldots, \mathbf{x}_t^{j_k})$, i.e., the tuple of the states of the object parts in $J$. For instance, on Fig. 2, if $J = \{2, 3\}$, then $\mathbf{x}_t^J$ represents the state of the whole left arm. Similarly, let $\mathbf{x}_t^{(i),J}$ denote the tuple of the parts in $J$ of the $i$th particle. For instance, for $J = \{2, 3\}$, $\mathbf{x}_t^{(i),J}$ corresponds the state of left arm as represented by the $i$th particle. In the rest of the paper, we will assume that the object is composed of precisely $P$ parts (in Fig. 2, $P = 6$). Now, we shall describe a slight generalization of PS where PF is iteratively applied on sets of object parts instead of just singletons like PS does. When PF is applied on a set, it is applied independently on all its elements. We need to distinguish at each step of such tracking algorithm the parts that were already processed by PF from those that were not yet. Thus, for any step $j$,

- let $P_j$ denote the set of object parts being processed at the $j$th step (in the case of PS, $P_j = \{j\}$);
- let $Q_j = \sum_{h=1}^{j} P_h$ denote the set of all the object parts processed up to (including) the $j$th step;
- let $R_j = \sum_{h=j+1}^{P} P_h$ denote the set of the object parts yet to process after the $j$th step is completed.

Fig. 2 illustrates these notations: here, $P_1 = \{1\}$, i.e., PF is first applied only on the torso; $P_2 = \{2, 4, 6\}$, i.e., at its 2nd step, the tracking algorithm applies PF in parallel on parts 2, 4 and 6. Therefore, at the 2nd step, parts $Q_2 = \{1, 2, 4, 6\}$ have been processed and there remains to process parts $R_2 = \{3, 5\}$. Thus, if PF has propagated all the parts in $Q_2$ from time $t - 1$ to $t$, in the particles, the parts in $R_2$ still refer to time $t - 1$ (see Fig. 2). Let $K$ denote the number of steps of the tracking algorithm, i.e., the number of sets $P_j$ (for PS, $K = P$). To simplify the proofs in the rest of the paper, we shall fix $Q_0 = R_K = \emptyset$. Now, PS can be described in Algorithm 1. In [MacCormick and Isard, 2000], it is showed that this algorithm is mathematically sound when its resampling method is a *weighted resampling* using a $g$ function corresponding to $p(\mathbf{y}_t | \mathbf{x}_t)$ (see the next section).

## 3 RESAMPLING METHODS

Several resampling schemes are classically used, that we shall review briefly now. Comparisons of their pros and cons can be found in [Douc et al., 2005].

*Multinomial resampling* consists of selecting $N$ numbers $k_i$, $i = 1, \ldots, N$, w.r.t. a uniform distribution $U((0, 1])$ on $(0, 1]$. Then, sample $\mathcal{S} = \{\mathbf{x}_t^{(i)}, w_t^{(i)}\}$ is substituted by a new sample $\mathcal{S}' = \{\mathbf{x}_t^{(D(k_i))}, \frac{1}{N}\}$ where $D(k_i)$ is the unique integer $j$ such that $\sum_{h=1}^{j-1} w_t^{(h)} < k_i \leq \sum_{h=1}^{j} w_t^{(h)}$. If $(n_1, \ldots, n_N)$ denote the number of times each of the particles in $\mathcal{S}$ are duplicated, then $(n_1, \ldots, n_N)$ is distributed w.r.t. the multinomial distribution $\text{Mult}(N; w_t^{(1)}, \ldots, w_t^{(N)})$. *Stratified resampling* differs from multinomial resampling by selecting randomly the $k_i$'s w.r.t. the uniform distribution $U((\frac{i-1}{N}, \frac{i}{N}])$. In *systematic resampling*, some number $k$ is drawn w.r.t. $U((0, \frac{1}{N}])$ and, then, the $k_i$'s are defined as $k_i = \frac{i-1}{N} + k$.

*Residual resampling* [Liu and Chen, 1998] is a method very efficient for decreasing the variance of the particle set induced by the resampling step. It is performed in two steps. First, for every $i \in \{1, \ldots, N\}$, $n_i' = \lfloor N w_t^{(i)} \rfloor$ duplicates of particle $\mathbf{x}_t^{(i)}$ of $\mathcal{S}$ are inserted into $\mathcal{S}'$. The $N - \sum_{i=1}^{n} n_i'$ particles still needed to complete the $N$-sample $\mathcal{S}'$ are drawn randomly using the multinomial distribution $\text{Mult}(N - \sum_{i=1}^{n} n_i'; N w_t^{(1)} - n_1', \ldots, N w_t^{(N)} - n_N')$, for instance using the multinomial resampling algorithm. The weights assigned to all the particles in $\mathcal{S}'$ are $1/N$.

Finally, *weighted resampling* is defined as follows: let $g : \mathcal{X} \mapsto \mathbb{R}$ be any strictly positive continuous function, where $\mathcal{X}$ denotes the state space. Weighted resampling proceeds as follows: let $\rho_t$ be defined as $\rho_t(i) = g(\mathbf{x}_t^{(i)}) / \sum_{j=1}^{N} g(\mathbf{x}_t^{(j)})$ for $i = 1, \ldots, N$. Select independently indices $k_1, \ldots, k_N$ according to probability $\rho_t$. Finally, construct the new set of particles $\mathcal{S}' = \{\mathbf{x}_t^{(k_i)}, w_t^{(k_i)} / \rho_t(k_i)\}_{i=1}^{N}$. [MacCormick, 2000] shows that $\mathcal{S}'$ represents the same probability distribution as $\mathcal{S}$ while

**Input**: A particle set $\{\mathbf{x}_{t-1}^{(i)}, w_{t-1}^{(i)}\}$ at time $t - 1$, an image $\mathcal{I}$
**Output**: A particle set $\{\mathbf{x}_t^{(i)}, w_t^{(i)}\}$ at time $t$
$Q \leftarrow \emptyset$;   $R \leftarrow \{1, \ldots, P\}$
**for** $j = 1$ **to** $K$ **do**
  **foreach** $k$ **in** $P_j$ **do**
    $Q' \leftarrow Q \cup \{k\}$;   $R' \leftarrow R \setminus \{k\}$
    $\{(\mathbf{x}_t^{(i),Q'}, \mathbf{x}_{t-1}^{(i),R'})\} \leftarrow$ propagate $(\{\mathbf{x}_t^{(i),Q}, \mathbf{x}_{t-1}^{(i),R}\})$
    $\{(w_t^{(i),Q'}, w_{t-1}^{(i),R'})\} \leftarrow$
      correct $(\{(\mathbf{x}_t^{(i),Q'}, \mathbf{x}_{t-1}^{(i),R'}), (w_t^{(i),Q}, w_{t-1}^{(i),R})\}, \mathcal{I})$
    $Q \leftarrow Q'$;   $R \leftarrow R'$
  $\{(\mathbf{x}_t^{(i),Q}, \mathbf{x}_{t-1}^{(i),R}), (w_t^{(i),Q}, w_{t-1}^{(i),R})\} \leftarrow$
    resample $(\{(\mathbf{x}_t^{(i),Q}, \mathbf{x}_{t-1}^{(i),R}), (w_t^{(i),Q}, w_{t-1}^{(i),R})\})$
**return** $\{\mathbf{x}_t^{(i)}, w_t^{(i)}\}$

**Algorithm 1:** Partitioned Sampling PS.

focusing the particles on the peaks of $g$. Note however that, unlike the other resampling methods described above, weighted resampling does not assign equal weights ($1/N$) to all the particles. In the rest of the paper, we will need this "equal weight" feature, so whenever weighted resampling will be used, it will be implicitly followed by one of the other above resampling methods.

In the next section, we will propose a new resampling method that exploits the structure within articulated objects to improve the efficiency of particle filtering.

## 4 DBN-BASED COMBINATORIAL RESAMPLING

Our resampling scheme is suitable for particle filters as described in Algo. 1. More precisely, we will show in Subsection 4.1 that, in articulated object tracking, the set $\{1,\ldots,P\}$ of parts of the objects to track can be partitioned into some sets $\{P_1,\ldots,P_K\}$ such that those parts in each $P_j$ are all independent conditionally to $\cup_{h<j} P_h$. For instance, in Fig. 2, $P = 6$ and $K = 3$, $P_1 = \{1\}$ corresponds to the torso, $P_2 = \{2,4,6\}$ to the head and both arms, and $P_3 = \{3,5\}$ to the forearms. In addition, given the position of the torso ($P_1$), those of the head and the arms ($P_2$) are independent. In Subsection 4.2, these independences will be exploited to justify that permutations of some particles' parts do not alter the estimation of $p(\mathbf{x}_{1:t}|\mathbf{y}_{1:t})$. Then, our resampling scheme, which will be described in Subsection 4.3, will exploit these permutations to construct implicitly some new exponential-size sample from which it will resample new high-quality samples.

### 4.1 IDENTIFYING SETS $P_1,\ldots,P_K$

To be sound, i.e., to not alter the estimation of $p(\mathbf{x}_{1:t}|\mathbf{y}_{1:t})$, Combinatorial Resampling exploits conditional independences among the different parts of the object. The partition into sets $P_1,\ldots,P_K$ precisely accounts for these independences and thus naturally results from a $d$-separation analysis, the independence property at the core of DBNs:

**Definition 1 ($d$-separation [Pearl, 1988])** *Two nodes $\mathbf{x}_t^i$ and $\mathbf{x}_s^j$ of a DBN are dependent conditionally to a set of nodes $\mathbf{Z}$ if and only if there exists a chain, i.e., an undirected path, $\{\mathbf{c}_1 = \mathbf{x}_t^i,\ldots,\mathbf{c}_n = \mathbf{x}_s^j\}$ linking $\mathbf{x}_t^i$ and $\mathbf{x}_s^j$ in the DBN such that the following two conditions hold:*

1. *for every node $\mathbf{c}_k$ such that the arcs are $\mathbf{c}_{k-1} \to \mathbf{c}_k \leftarrow \mathbf{c}_{k+1}$, either $\mathbf{c}_k$ or one of its descendants is in $\mathbf{Z}$;*

2. *none of the other nodes $\mathbf{c}_k$ belongs to $\mathbf{Z}$.*

*Such a chain is called* active *(else it is* blocked*). If there exists an active chain linking two nodes, these nodes are dependent and are called $d$-connected, otherwise they are independent conditionally to $\mathbf{Z}$ and are called $d$-separated.*

In Fig. 2, conditionally to the position of the torso up to time $t$, both arms are thus independent.

In the rest of the paper, we will assume that, *within each time slice*, the DBN structure is a directed tree, i.e., there do not exist nodes $\mathbf{x}_t^i, \mathbf{x}_t^j, \mathbf{x}_t^k$ such that $\mathbf{x}_t^i \to \mathbf{x}_t^j \leftarrow \mathbf{x}_t^k$. We will also assume that arcs across time slices link similar nodes, i.e., there exist no arc $\mathbf{x}_{t-1}^i \to \mathbf{x}_t^j$ with $j \neq i$. Finally, we will assume that nodes $\mathbf{y}_t^j$ have only one parent $\mathbf{x}_t^j$ and no children. For articulated object tracking, these requirements are rather mild and Fig. 2 satisfies all of them.

Now, we can construct sets $P_1,\ldots,P_K$: for any node, say $X_t$, in time slice $t$ of the DBN, let $\mathbf{Pa}(X_t)$ and $\mathbf{Pa}_t(X_t)$ denote the set of parents of $X_t$ in the DBN in all time slices and in time slice $t$ only respectively. For instance, in Fig. 2, $\mathbf{Pa}(\mathbf{x}_t^2) = \{\mathbf{x}_t^1, \mathbf{x}_{t-1}^2\}$ and $\mathbf{Pa}_t(\mathbf{x}_t^2) = \{\mathbf{x}_t^1\}$. Let $\{P_1,\ldots,P_K\}$ be a partition of $\{1,\ldots,P\}$ defined by:

- $P_1 = \{k \in \{1,\ldots,P\} : \mathbf{Pa}_t(\mathbf{x}_t^k) = \emptyset\}$;

- for any $j > 1$, $P_j = \{k \in \{1,\ldots,P\}\setminus\bigcup_{h=1}^{j-1} P_h : \mathbf{Pa}_t(\mathbf{x}_t^k) \subseteq \bigcup_{h=1}^{j-1}\bigcup_{r\in P_h}\{\mathbf{x}_t^r\}\}$.

On Fig. 2, this results in $P_1 = \{1\}$, $P_2 = \{2,4,6\}$ and $P_3 = \{3,5\}$. It turns out that the way we constructed the $P_j$'s, all the $\mathbf{x}_t^k \in P_j$ can be processed independently by PF because they are independent conditionally to $\mathbf{Pa}(\mathbf{x}_t^k)$, and this is precisely this independence property which is needed to enable a sound object part swapping within Combinatorial Resampling:

**Proposition 1** *The particle set resulting from Algorithm 1, with $P_j$ defined as in the preceding paragraph, $Q_j = \sum_{h=1}^{j} P_h$ and $R_j = \sum_{h=j+1}^{K} P_h$, represents $p(\mathbf{x}_t|\mathbf{y}_{1:t})$.*

**Proof:** By induction on $j$. Assume that, before processing parts $P_j$, particles estimate $p(\mathbf{x}_t^{Q_{j-1}}, \mathbf{x}_{t-1}^{R_{j-1}}|\mathbf{y}_{1:t-1}, \mathbf{y}_t^{Q_{j-1}})$. This is clearly the case for $P_1$ since $P_1$ are the first parts processed. Remember that $P_j, Q_j, R_j$ are the set of parts processed at the $j$th step, up to the $j$th step and still to process respectively. We will now examine sequentially the distributions estimated by the particle set after applying in parallel PF's prediction step over the parts in $P_j$, then after applying PF's correction step and, finally, after resampling.

1. Let us show that after the parallel propagations of the parts in $P_j$ (prediction step), the particle set represents density $p(\mathbf{x}_t^{Q_j}, \mathbf{x}_{t-1}^{R_j}|\mathbf{y}_{1:t-1}, \mathbf{y}_t^{Q_{j-1}})$. For instance, on Fig. 2, this means that, after propagating the parts in $P_2$, the particle set estimates $p(\mathbf{x}_t^{1,2,4,6}, \mathbf{x}_{t-1}^{3,5}|\mathbf{y}_{1:t-1}, \mathbf{y}_t^1)$, i.e., only the positions of the forearms still refer to time $t-1$ and the only observation taken into account at time $t$ is the position of the torso (not yet those of the head and arms). All these

parallel operations correspond to computing:

$$\int p(\mathbf{x}_t^{Q_{j-1}}, \mathbf{x}_{t-1}^{R_{j-1}}|\mathbf{y}_{1:t-1}, \mathbf{y}_t^{Q_{j-1}}) \prod_{k \in P_j} p(\mathbf{x}_t^k|\mathbf{Pa}(\mathbf{x}_t^k)) \, d\mathbf{x}_{t-1}^{P_j}.$$

By $d$-separation, a node is independent of all of its non descendants conditionally to its parents. Hence, for every $k \in P_j$, $\mathbf{x}_t^k$ is independent of $\mathbf{x}_t^{P_j \setminus \{k\}} \cup \mathbf{x}_t^{Q_{j-1}} \cup \mathbf{x}_{t-1}^{R_{j-1}} \cup \mathbf{y}_{1:t-1} \cup \mathbf{y}_t^{Q_{j-1}}$ conditionally to $\mathbf{Pa}(\mathbf{x}_t^k)$ (see Fig. 3.a where $\mathbf{x}_t^k$ is the doubly-circled node, $\mathbf{Pa}(\mathbf{x}_t^k)$ are the striped nodes and the black and shaded nodes correspond to the independent observation and state nodes respectively). Consequently, the above integral is equivalent to:

$$\int p(\mathbf{x}_t^{Q_{j-1}}, \mathbf{x}_{t-1}^{R_{j-1}}|\mathbf{y}_{1:t-1}, \mathbf{y}_t^{Q_{j-1}})$$
$$p(\mathbf{x}_t^{P_j}|\mathbf{x}_t^{Q_{j-1}}, \mathbf{x}_{t-1}^{R_{j-1}}, \mathbf{y}_{1:t-1}, \mathbf{y}_t^{Q_{j-1}}) \, d\mathbf{x}_{t-1}^{P_j}$$
$$= \int p(\mathbf{x}_t^{P_j}, \mathbf{x}_t^{Q_{j-1}}, \mathbf{x}_{t-1}^{R_{j-1}}|\mathbf{y}_{1:t-1}, \mathbf{y}_t^{Q_{j-1}}) \, d\mathbf{x}_{t-1}^{P_j}.$$

As $Q_j = Q_{j-1} \cup P_j$ and $R_{j-1} = P_j \cup R_j$, the above equation is equivalent to $p(\mathbf{x}_t^{Q_j}, \mathbf{x}_{t-1}^{R_j}|\mathbf{y}_{1:t-1}, \mathbf{y}_t^{Q_{j-1}})$.

2. Let us show that after the parallel corrections of the $P_j$ parts, the particle set estimates $p(\mathbf{x}_t^{Q_j}, \mathbf{x}_{t-1}^{R_j}|\mathbf{y}_{1:t-1}, \mathbf{y}_t^{Q_j})$. These operations correspond to computing, up to a constant, distribution $p(\mathbf{x}_t^{Q_j}, \mathbf{x}_{t-1}^{R_j}|\mathbf{y}_{1:t-1}, \mathbf{y}_t^{Q_{j-1}}) \times \prod_{k \in P_j} p(\mathbf{y}_t^k|\mathbf{x}_t^k)$. By $d$-separation, nodes $\mathbf{y}_t^k$ are independent of the rest of the DBN conditionally to $\mathbf{x}_t^k$, so $p(\mathbf{y}_t^{P_j}|\mathbf{x}_t^{Q_j}, \mathbf{x}_{t-1}^{R_j}, \mathbf{y}_{1:t-1}, \mathbf{y}_t^{Q_{j-1}}) = \prod_{k \in P_j} p(\mathbf{y}_t^k|\mathbf{x}_t^k)$. After the corrections over $P_j$, the particle set thus estimates $p(\mathbf{x}_t^{Q_j}, \mathbf{x}_{t-1}^{R_j}, \mathbf{y}_t^{P_j}|\mathbf{y}_{1:t-1}, \mathbf{y}_t^{Q_{j-1}})$, which, when normalized, is equal to $p(\mathbf{x}_t^{Q_j}, \mathbf{x}_{t-1}^{R_j}|\mathbf{y}_{1:t-1}, \mathbf{y}_t^{Q_{j-1}}, \mathbf{y}_t^{P_j}) = p(\mathbf{x}_t^{Q_j}, \mathbf{x}_{t-1}^{R_j}|\mathbf{y}_{1:t-1}, \mathbf{y}_t^{Q_j})$. As resamplings do not alter densities, at the end of the algorithm, the particle set estimates $p(\mathbf{x}_t^{Q_K}, \mathbf{x}_{t-1}^{R_K}|\mathbf{y}_{1:t-1}, \mathbf{y}_t^{Q_K}) = p(\mathbf{x}_t|\mathbf{y}_{1:t})$. □

### 4.2 SUBSTATE PERMUTATIONS

The advantage of using the $P_j$'s as defined above instead of singletons as the classical PS does is that this enables to improve by permutations the particle set without altering the joint posterior density. Those permutations are the core of Combinatorial Resampling as the latter creates implicitly new particle sets resulting from all the possible permutations. The next proposition determines the permutations that guarantee the distributions are not altered. Intuitively, it asserts that whenever two particles are such that they have the same states on some nodes $\mathbf{Pa}_s(\mathbf{x}_s^k)$, then swapping their states on $\mathbf{x}_s^k$ and its descendants cannot alter the density estimated by the particle set. For instance, on Fig. 3.b, if two particles have the same value for the striped nodes $\mathbf{Pa}_s(\mathbf{x}_s^k)$, their values on the shaded node $\mathbf{x}_s^k$ and the black one ($\mathbf{x}_s^k$'s descendant) can be safely swapped.

**Proposition 2** *Let $\{(\mathbf{x}_t^{(i),Q_j}, \mathbf{x}_{t-1}^{(i),R_j})\}$ be the particle set at the jth step of Algo. 1. Let $k \in P_j$ and let $\mathbf{Desc}_t(\mathbf{x}_t^k)$ be the set of descendants of $\mathbf{x}_t^k$ in time slice $t$. Let $\sigma : \{1, \ldots, N\} \mapsto \{1, \ldots, N\}$ be any permutation such that $\mathbf{x}_s^{(i),h} = \mathbf{x}_s^{(\sigma(i)),h}$ for all the nodes $\mathbf{x}_s^h \in \cup_{s=1}^t \mathbf{Pa}_s(\mathbf{x}_s^k)$. Then, the particle set resulting from the application of $\sigma$ on the parts of $\{(\mathbf{x}_t^{(i),Q_j}, \mathbf{x}_{t-1}^{(i),R_j})\}$ corresponding to $\{\mathbf{x}_t^k\} \cup \mathbf{Desc}_{t-1}(\mathbf{x}_{t-1}^k)$ still estimates $p(\mathbf{x}_t^{Q_j}, \mathbf{x}_{t-1}^{R_j}|\mathbf{y}_{1:t-1}, \mathbf{y}_t^{Q_j})$.*

**Proof:** If $j = 1$, the proposition trivially holds since $\sigma$ is applied to all the nodes of the connected component of $\mathbf{x}_t^k$. Assume now that $j \neq 1$. We shall now partition the object parts as described on Fig. 3.c to highlight which parts shall be permuted, which ones shall be identical to enable permutations and which parts are unconcerned: let $\mathbf{x}_{t-1}^D = \mathbf{Desc}_{t-1}(\mathbf{x}_{t-1}^k)$, $\mathbf{x}_t^{k'} = \mathbf{Pa}_t(\mathbf{x}_t^k)$, $\mathbf{x}_t^V = \mathbf{x}_t^{Q_j} \setminus (\{\mathbf{x}_t^k, \mathbf{x}_t^{k'}\})$ and $\mathbf{x}_{t-1}^W = \mathbf{x}_{t-1}^{R_j} \setminus \mathbf{x}_{t-1}^D$. Thus, the permuted parts are $\mathbf{x}_t^k \cup \mathbf{x}_{t-1}^D$ (see Fig. 3.c), the identical part is $\mathbf{x}_t^{k'}$, and the unconcerned parts are $\mathbf{x}_t^V \cup \mathbf{x}_{t-1}^W$.

$$p(\mathbf{x}_t^{Q_j}, \mathbf{x}_{t-1}^{R_j}|\mathbf{y}_{1:t-1}, \mathbf{y}_t^{Q_j}) \propto p(\mathbf{x}_t^{Q_j}, \mathbf{x}_{t-1}^{R_j}, \mathbf{y}_{1:t-1}, \mathbf{y}_t^{Q_j})$$
$$= p(\mathbf{x}_t^{\{k,k'\} \cup V}, \mathbf{x}_{t-1}^{D \cup W}, \mathbf{y}_{1:t}^{\{k,k'\} \cup V}, \mathbf{y}_{1:t-1}^{D \cup W})$$
$$= \int p(\mathbf{x}_t^{\{k\} \cup V}, \mathbf{x}_{1:t}^{k'}, \mathbf{x}_{t-1}^{D \cup W}, \mathbf{y}_{1:t}^{\{k,k'\} \cup V}, \mathbf{y}_{1:t-1}^{D \cup W}) d\mathbf{x}_{1:t-1}^{k'}$$

Conditionally to $\{\mathbf{x}_{1:t}^{k'}\}$, $S = \{\mathbf{x}_t^k\} \cup \mathbf{x}_{t-1}^D \cup \mathbf{y}_{1:t}^k \cup \mathbf{y}_{1:t-1}^D$ is independent of the rest of the DBN because, by Definition 1, no active chain can pass through an arc outgoing from a node in a conditioning set and, removing from the DBN the arcs outgoing from $\{\mathbf{x}_{1:t}^{k'}\}$, $S$ is not connected anymore to the rest of the DBN. For the same reason, $\mathbf{x}_t^V \cup \mathbf{x}_{t-1}^W \cup \mathbf{y}_{1:t}^V \cup \mathbf{y}_{1:t-1}^W$ is independent of the rest of the DBN conditionally to $\{\mathbf{x}_{1:t}^{k'}\}$. Therefore, the above integral is equal to:

$$\int p(\mathbf{x}_{1:t}^{k'}, \mathbf{y}_{1:t}^{k'}) \, p(\mathbf{x}_t^k, \mathbf{x}_{t-1}^D, \mathbf{y}_{1:t}^k, \mathbf{y}_{1:t-1}^D | \mathbf{x}_{1:t}^{k'}) \\ p(\mathbf{x}_t^V, \mathbf{x}_{t-1}^W, \mathbf{y}_{1:t}^V, \mathbf{y}_{1:t-1}^W | \mathbf{x}_{1:t}^{k'}) \, d\mathbf{x}_{1:t-1}^{k'}. \quad (3)$$

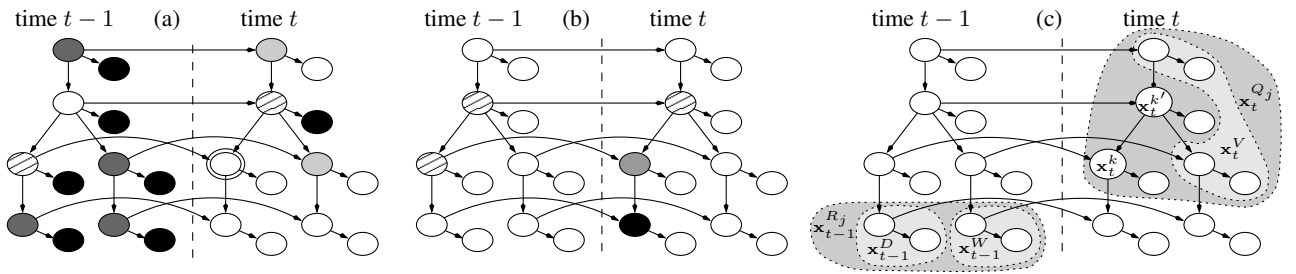

Figure 3: $d$-separation analysis.

Permuting particles over parts $\{\mathbf{x}_t^k\} \cup \mathbf{x}_{t-1}^D$ for fixed values of $\mathbf{x}_{1:t}^{k'}$ cannot change the estimation of density $p(\mathbf{x}_t^k, \mathbf{x}_{t-1}^D, \mathbf{y}_{1:t}^k, \mathbf{y}_{1:t-1}^D | \mathbf{x}_{1:t}^{k'})$ because estimations by samples are insensitive to the order of the elements in the samples. Moreover, it can neither affect the estimation of density $p(\mathbf{x}_t^V, \mathbf{x}_{t-1}^W, \mathbf{y}_{1:t}^V, \mathbf{y}_{1:t-1}^W | \mathbf{x}_{1:t}^{k'})$ since $\mathbf{x}_t^V \cup \mathbf{x}_{t-1}^W \cup \mathbf{y}_{1:t}^V \cup \mathbf{y}_{1:t-1}^W$ is independent of $\{\mathbf{x}_t^k\} \cup \mathbf{x}_{t-1}^D$ conditionally to $\{\mathbf{x}_{1:t}^{k'}\}$. Consequently, applying permutation $\sigma$ on parts $\{\mathbf{x}_t^k\} \cup \mathbf{x}_{t-1}^D$ does not change the estimation of Eq. (3) and, therefore, of $p(\mathbf{x}_t^{Q_j}, \mathbf{x}_{t-1}^{R_j} | \mathbf{y}_{1:t-1}, \mathbf{y}_t^{Q_j})$. □

As shown in the next subsection, these permutations can be exploited at the resampling level to improve samples.

### 4.3 OUR RESAMPLING APPROACH

All the permutations satisfying Proposition 2 can be applied to the particle set without altering the estimation of the posterior density. For instance, let $\mathbf{x}_t^{(1)}$ and $\mathbf{x}_t^{(2)}$ be two particles whose torso positions are identical, then swapping their left arm and forearm positions $(\mathbf{x}_t^2, \mathbf{x}_{t-1}^3)$ cannot alter the density estimation. Similarly, the latter is unaffected by duplications of all the particles within a particle set. This leads to Combinatorial Resampling:

**Definition 2 (Combinatorial Resampling)** *Let $S$ be the particle set at the jth step of Algo. 1. For any $k \in P_j$, let $\Sigma_k$ be the set of permutations satisfying Proposition 2. Let $\Sigma = \prod_{k \in P_j} \Sigma_k$. Let $S' = \cup_{\sigma \in \Sigma} \{particle\ set\ resulting\ from\ the\ application\ of\ \sigma\ to\ S\}$. Combinatorial resampling consists of applying any resampling algorithm over the combinatorial set $S'$ instead of $S$.*

On the example of Fig. 2, let $\mathbf{x}_t^{(1)} = \langle 1, 2, 3, 4, 5, 6 \rangle$, $\mathbf{x}_t^{(2)} = \langle 1, 2', 3', 4', 5', 6' \rangle$ and $\mathbf{x}_t^{(3)} = \langle 1'', 2'', 3'', 4'', 5'', 6'' \rangle$ be three particles, where each number, $1, 1'', 2, 2', 2''$, etc., corresponds to the state of a part in Fig. 2. Assume that $S = \{\mathbf{x}_t^{(1)}, \mathbf{x}_t^{(2)}, \mathbf{x}_t^{(3)}\}$ at the 2nd step of Algo. 1, i.e., the object parts just processed are $P_2 = \{2, 4, 6\}$. Parts $\{2, 3\}$, $\{4, 5\}$ and $\{6\}$ can be permuted in $\mathbf{x}_t^{(1)}$ and $\mathbf{x}_t^{(2)}$ because their torso, i.e. 1, are identical, hence $S'$ is the union of the result of all such permutations over $S$ and is thus equal to:

$\langle 1,2\ ,3\ ,4\ ,5\ ,6\ \rangle\ \langle 1,2',3',4',5',6' \rangle\ \langle 1'',2'',3'',4'',5'',6'' \rangle$
$\langle 1,2\ ,3\ ,4\ ,5\ ,6'\rangle\ \langle 1,2',3',4',5',6\ \rangle\ \langle 1'',2'',3'',4'',5'',6'' \rangle$
$\langle 1,2\ ,3\ ,4',5',6\ \rangle\ \langle 1,2',3',4\ ,5\ ,6'\rangle\ \langle 1'',2'',3'',4'',5'',6'' \rangle$
$\langle 1,2\ ,3\ ,4',5',6'\rangle\ \langle 1,2',3',4\ ,5\ ,6\ \rangle\ \langle 1'',2'',3'',4'',5'',6'' \rangle$
$\langle 1,2',3',4\ ,5\ ,6\ \rangle\ \langle 1,2\ ,3\ ,4',5',6'\rangle\ \langle 1'',2'',3'',4'',5'',6'' \rangle$
$\langle 1,2',3',4\ ,5\ ,6'\rangle\ \langle 1,2\ ,3\ ,4',5',6\ \rangle\ \langle 1'',2'',3'',4'',5'',6'' \rangle$
$\langle 1,2',3',4',5',6\ \rangle\ \langle 1,2\ ,3\ ,4\ ,5\ ,6'\rangle\ \langle 1'',2'',3'',4'',5'',6'' \rangle$
$\langle 1,2',3',4',5',6'\rangle\ \langle 1,2\ ,3\ ,4\ ,5\ ,6\ \rangle\ \langle 1'',2'',3'',4'',5'',6'' \rangle$

Constructing $S'$ in extension is impossible in practice because $|\Sigma|$ grows exponentially with $N$, the number of particles. Fortunately, we can sample over $S'$ without actually constructing it. We shall explain the idea on the particle set $S$ illustrated on Fig. 4, which corresponds to the object of Fig. 2 in which we omitted the head part for clarity reasons. Assume that parts $P_j = \{3, 5\}$, i.e., the forearms, have just been processed and we wish to sample over combinatorial sample $S'$ induced by $S$. To construct a new particle, the idea is to first select a value for the parts in $Q_{j-1}$, i.e., those processed at previous steps by PF and in which no permutation will occur. Here, $Q_{j-1} = \{1, 2, 4\}$. We thus first determine the different values of $\mathbf{x}_t^{Q_{j-1}}$ in $S$ and partition $S$ into sets $S_1, \ldots, S_R$ such that all the particles in each set $S_h$ have the same value for $\mathbf{x}_t^{Q_{j-1}}$ (see Fig. 4). In this figure, $S_1$ thus contains the first two particles since their values on $\mathbf{x}_t^{Q_{j-1}}$ are both $\langle 1, 0, 3 \rangle$. To each such set $S_h$ is assigned a weight $W_h$ defined below so that picking the value of $\mathbf{x}_t^{Q_{j-1}}$ in $S_h$ w.r.t. weight $W_h$ results in a particle set estimating the same distribution as that of $S$. Once the value of $\mathbf{x}_t^{Q_{j-1}}$ has been chosen, there just remains to pick independently values for each $\mathbf{x}_t^k$ and their descendants, $k \in P_j$, that are compatible with that chosen for $\mathbf{x}_t^{Q_{j-1}}$. Thus, for any $h \in \{1, \ldots, R\}$, let $S_h^k$ denote the set of particles in $S$ whose $k$th part value is compatible with the value of $\mathbf{x}_t^{Q_{j-1}}$ in $S_h$. By Proposition 2, $S_h^k$ is the set of particles in $S$ that have the same value of $\mathbf{Pa}_t(\mathbf{x}_t^k)$ as those in $S_h$. For instance, in Fig. 4, $S_1^3$ is the set of the first 3 particles because all of them have value 1 on part 2.

Now, to determine the aforementioned weights $W_h$, there just needs to count how many times the combinatorial set has duplicated $S_h$. So, let $N_1, \ldots, N_R$ and $N_1^k, \ldots, N_R^k$ denote the sizes of $S_1, \ldots, S_R$ and $S_1^k, \ldots, S_R^k$ respectively. Finally, let $N^k = \max\{N_1^k, \ldots, N_R^k\}$ and, for any $h \in \{1, \ldots, R\}$, let $W_h^k$ denote the sum of the weights assigned to the $k$th part of the particles in $S_h^k$, i.e., $W_h^k = \sum_{\mathbf{x}_t^{(i)} \in S_h^k} w^{(i),k}$. Then, as proved below, for any $h$,

parts:
$$S_1^3 = S_2^3 \begin{pmatrix} \begin{array}{ccccc} 3 & 2 & 1 & 4 & 5 \\ \hline 5 & 1 & 0 & 3 & 8 \\ 6 & 1 & 0 & 3 & 9 \\ 7 & 1 & 0 & 4 & 12 \\ 10 & 2 & 0 & 4 & 13 \\ 11 & 2 & 0 & 4 & 14 \end{array} \end{pmatrix} \begin{matrix} S_1 \\ \\ S_2 \\ S_3 \end{matrix}$$

$P_j = \{\text{parts 3,5}\}$
$Q_{j-1} = \{\text{parts 1,2,4}\}$
$\mathbf{Pa}_t(\mathbf{x}_t^3) = \{\mathbf{x}_t^2\}$
$\mathbf{Pa}_t(\mathbf{x}_t^5) = \{\mathbf{x}_t^4\}$

Figure 4: Sets $S_{\mathbf{i}_h}$ and $S_{\mathbf{i}_h^k}$. Each row represents a particle $\mathbf{x}_t^{(i)}$ and each number a value $\mathbf{x}_t^{(i),j}$ of part $j$ of the particle.

**Input**: A particle set $\{(\mathbf{x}_t^{(i),Q_j}, \mathbf{x}_{t-1}^{(i),R_j}), w_t^{(i)}\}_{i=1}^N$
**Output**: A new particle set $\{(\mathbf{x}_t''^{(i),Q_j}, \mathbf{x}_{t-1}''^{(i),R_j}), w_t''^{(i)}\}_{i=1}^N$

1 **for** $i = 1$ **to** $N$ **do**
2    $h \leftarrow$ sample $\{1, \ldots, R\}$ w.r.t. weights $W_1, \ldots, W_R$
3    $\mathbf{x}_t''^{(i),Q_{j-1}} \leftarrow \mathbf{x}_t^{(z),Q_{j-1}}$ where $\mathbf{x}_t^{(z)}$ is any element in $S_h$
4    $w_t''^{(i)} \leftarrow 1$
5    **foreach** $k$ **in** $P_j$ **do**
6      $\mathbf{x}_t^{(r)} \leftarrow$ sample from $S_h^k$ w.r.t. weights $\{w_t^{(r),k}\}_{\mathbf{x}_t^r \in S_h^k}$
7      $\mathbf{x}_t''^{(i),k} \leftarrow \mathbf{x}_t^{(r),k}$;   $w_t''^{(i)} \leftarrow w_t''^{(i)} \times w_t^{(r),k}$
8      $\mathbf{x}_{t-1}''^{(i),\mathbf{Desc}_{t-1}(\mathbf{x}_t^k)} \leftarrow \mathbf{x}_{t-1}^{(r),\mathbf{Desc}_{t-1}(\mathbf{x}_t^k)}$

9 **return** $\{\mathbf{x}_t''^{(i)}, w_t''^{(i)}\}_{i=1}^N$

**Algorithm 2:** Efficient resampling over $S'$.

$$W_h = N_h \times \prod_{k \in P_j} \frac{N^k!}{A_{N_h^k}^{N_h}} \times A_{N_h^k - 1}^{N_h - 1} \times W_h^k, \quad (4)$$

where $A_n^k = n!/(n-k)!$ stands for the number of $k$-permutations out of $n$ elements. Resampling over $S'$ can thus be performed efficiently as in Algo. 2. To scale-up to large particle sets, $\log(W_h)$ should be computed instead of $W_h$ and the weights used in line 2 of Algo. 2 should be $\exp(\log W_h - \log W)$, where $W = \max\{W_1, \ldots, W_R\}$.

**Proposition 3** *Algorithm 2 produces a particle set estimating the same density as that given in input.*

**Proof:** Let $S = \{(\mathbf{x}_t^{(i),Q_j}, \mathbf{x}_{t-1}^{(i),R_j})\}_{i=1}^N$ and $S'$ its combinatorial set (see Def. 2). In lines 2–3, Algo. 2 selects which central part $Q_{j-1}$ particle $\mathbf{x}_t''$ should have. By definition, this amounts to selecting one set $S_h$ w.r.t. the sum of the weights of the particles in $S'$ having the same central part as those in $S_h$. Let us show that this is achieved using the weights described in Eq. (4).

In Definition 2, $\Sigma_k$ is the set of all the possible permutations of the $k$th part of the particles in $S$. Clearly, within each set $S_h^k$, all the $N_h^k!$ permutations of the $k$th part of the particles of this set are admissible. They form the cycles within the permutations of $\Sigma_k$ and, as such, a given permutation $\sigma$ over $S_h^k$ shall appear many times within $\Sigma_k$. There is no need to count precisely how many times $\sigma$ is repeated, what is important is that the repeated sets of particles estimate the same density as $S$. To do so, remark that $N^k = \max\{N_1^k, \ldots, N_R^k\}$ is the size of the biggest set $S_1^k, \ldots, S_R^k$. Applying all the permutations over this set multiplies its size by $N^k!$, so the size of all the other sets should be multiplied by the same amount. Duplicating $N^k!/N_h^k!$ permutation $\sigma$ guarantees that all the $Q_{j-1}$-central parts of the particles in $S$ are duplicated the same number of times. Now, the particles in $S_h$ also belong to $S_h^k$. As $|S_h| = N_h$, there are $A_{N_h^k}^{N_h}$ different possibilities to assign some $k$-part of $S_h^k$ to the particles of $S_h$. The number of times these permutations are repeated within those over $S_h^k$ is thus $N_h^k!/A_{N_h^k}^{N_h}$. Hence, duplicating $(N^k!/N_h^k!) \times (N_h^k!/A_{N_h^k}^{N_h}) = N^k!/A_{N_h^k}^{N_h}$ times the permutations over $S_h$ ensures that the particle set estimates the same density as $S$. The same applies to all the other parts in $P_j$, hence the product in Eq. (4).

Now, let us compute the sum of the weights of the particles resulting from all the permutations over $S_h$. Each such permutation generates a new set of $N_h$ particles. By symmetry, if $\mathbf{W}$ is the sum of the weights of the first particle in each set, call it $\mathbf{x}_t^{(i)}$, then the overall sum we look for is $N_h \times \mathbf{W}$. As permutations over the parts in $P_j$ are independent, $\mathbf{W}$ is equal to the product over parts $k \in P_j$ of the sum $\mathbf{W}^k$ of the weights induced by all the permutations over the $k$th part, i.e., the permutations over $S_h^k$. By symmetry, any weight in $S_h^k$ can be assigned to $\mathbf{x}_t^{(i)}$, hence $\mathbf{W}^k$ is equal to the sum of all these weights, $W_h^k$, times the number $\mathbf{O}$ of occurrences of each weight induced by all the permutations over $S_h^k$. For instance, if there are 3 weights 1,2,3, then there are $\mathbf{O} = 2$ permutations where the first particle as a weight of 1: $\langle 1, 2, 3 \rangle$ and $\langle 1, 3, 2 \rangle$. Once particle $\mathbf{x}_t^{(i)}$ has been assigned a weight, there remains $N_h - 1$ weights to assign to the other particles from a set of $N_h^k - 1$ weights, hence there are $\mathbf{O} = A_{N_h^k - 1}^{N_h - 1}$ possibilities. Overall, $\mathbf{W}^k$ is thus equal to $W_h^k \times A_{N_h^k - 1}^{N_h - 1}$ and we get Eq. (4).

So, lines 2–3 select correctly the $Q_{j-1}$ part. Once this is done, by $d$-separation, all the parts in $P_j$ are independent and should be sampled w.r.t. $p(\mathbf{x}_t^k | \mathbf{Pa}_t(\mathbf{x}_t^k))$, which is done in lines 5–8 since $p(\mathbf{x}_t^k | \mathbf{Pa}_t(\mathbf{x}_t^k)) \propto w_t^{(i),k}$.   □

We shall now provide some experiments highlighting the efficiency of our resampling scheme.

## 5 EXPERIMENTATIONS

We performed experiments on synthetic data in order to create sequences varying the criteria whose impact on our algorithm's efficiency are the most important, i.e. the number of parts processed in parallel and the length of the object's arms. As such, this resulted in a fine picture of the behaviors of our algorithm. These results are given in Subsection 5.1. Of course, our algorithm is also effective on real sequences. This is illustrated in Subsection 5.2.

For both cases, articulated objects are modeled by a set of $P$ polygonal parts (or regions): a central one $P_1$ (containing only one polygon) to which are linked $|P_j|$, $j > 1$, arms of length $K - 1$ (see Fig. 5 for some examples). The polygons are manually positioned in the first frame. State vectors contain the parameters describing all the parts and are defined by $\mathbf{x}_t = \{\mathbf{x}_t^1, \ldots, \mathbf{x}_t^P\}$, with $\mathbf{x}_t^k = \{x_t^k, y_t^k, \theta_t^k\}$, where $(x_t^k, y_t^k)$ is the center of part $k$, and $\theta_t^k$ is its orientation, $k = 1, \ldots, P$. We thus have $|\mathcal{X}| = 3P$. A particle $\mathbf{x}_t^{(i)} = \{\mathbf{x}_t^{(i),1}, \ldots, \mathbf{x}_t^{(i),P}\}$, $i = 1, \ldots, N$, is a possible spatial configuration, i.e., a realization, of the ar-ticulated object. Particles are propagated using a random walk whose variances $\sigma_x$, $\sigma_y$ and $\sigma_\theta$ have been empirically fixed. Particle weights are computed by measuring the similarity between the distribution of pixels in the region of the estimated part of the object and that of the corresponding reference region using the Bhattacharyya distance [Bhattacharyya, 1943]. The particle weights are then computed by $w_{t+1}^{(i)} = w_t^{(i)} p(\mathbf{y}_{t+1}|\mathbf{x}_{t+1}^{(i)}) \propto w_t^{(i)} e^{-\lambda \mathbf{d}^2}$, with, in our tests, $\lambda = 50$ and $\mathbf{d}$ the Bhattacharyya distance between the target (prior) and the reference (previously estimated) 8-bin histograms. The articulated object's distribution is estimated starting from its central part $P_1$.

### 5.1 TESTS ON SYNTHETIC VIDEO SEQUENCES

We have generated our own video sequences composed of 300 frames of $800 \times 640$ pixels. Each video displays an articulated object randomly moving and deforming over time, subject to either weak or strong motions. Some examples are given in Fig. 5. With various numbers of parts, the articulated objects are designed to test the ability of resampling to deal with high-dimensional state spaces.

We compare six different resampling approaches. The first five (multinomial, systematic, stratified, residual and weighted resampling) are integrated into PS. PS propagates and corrects particles polygon after polygon to derive a global estimation of the object. For combinatorial resampling, the object's arms are considered independent conditionally to the central part and, thus, the $P_j$ parts, $j > 1$, correspond to the $j$th joints of all the arms. For weighted resampling, function $g$ is set empirically to $g(w) = e^{20w}$ to favor the selection of high-weighted particles over low-weighted ones. Results are compared w.r.t. two criteria: computation times and estimation errors, defined as the sum of the Euclidean distances between each corner of the estimated parts and its sibling in the ground truth. For all these tests, we fixed $\sigma_x = \sigma_y = 1$ pixel and $\sigma_\theta = 0.025$ rad. All the results presented are a mean over 30 runs performed on a MacBook Pro with a 2.66 GHz Intel Core i7.

We first compared the estimation errors. Fig 6.(a-c) show a convergence study of the resampling methods depending on the number $N$ of particles for the 3 objects of Fig. 5. For all these tests, combinatorial resampling (CR) outperforms all the other methods: i) it converges faster (about only $N = 100$ particles are necessary to do so) when the other methods often require 300 particles to converge; ii) CR's error at convergence is much lower than that of the other methods. For instance, in Fig.6(a), CR reaches the convergence error of the other methods (about 230 pixels) with only $N = 20$ particles and, with 100 particles, its error decreases to 112 pixels. When the length of the arms (given by $K$) increases (Fig 6(b)), CR stays robust, whereas multinomial, systematic, stratified and residual resampling tend to fail (estimation errors twice higher). Weighted resampling seems more stable, but gives estimation errors 25% higher than those of CR. Finally, when the number of parts treated in parallel increases (Fig 6(c)), CR stays stable: with only $N = 20$ particles, its estimation error is 2.5 to 3 times lower than the one of other resampling approaches.

Finally, resampling times (in seconds) over the whole sequences, are reported in Table 1 for the estimation of the densities of the objects of Fig. 5(a-c) with $N = \{100, 600\}$ particles. The first four resampling approaches have similar behaviors. Due to its additional step ensuring that weights are equal to $1/N$, which is required by Algo. 1, weighted resampling is longer. The best approach is CR when the number of particles is high (600) and when the size of the $P_j$'s is high. For instance, when tracking the object of Fig. 5(c) (8 parts processed simultaneously), the resampling times are considerably lower with CR than with the other methods. This is due to the fact that, by processing several object parts simultaneously, the number of resamplings performed is significantly reduced. Hence, even if performing CR once is longer than performing another method, overall, CR is globally faster. Note also that CR's response times increase more slowly with $N$ than the other methods. Finally, when $K$ increases (Fig. 5(b)), our approach also provides significantly smaller resampling times when $N$ becomes high.

### 5.2 TESTS ON REAL VIDEO SEQUENCES

We tested our approach on sequences from the UCF50 dataset (http://server.cs.ucf.edu/~vision/data/UCF50.rar), to demonstrate the efficiency of our combinatorial resampling to make the particle set better focus on the modes

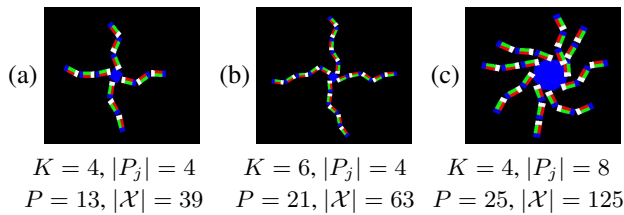

(a) $K = 4, |P_j| = 4$
$P = 13, |\mathcal{X}| = 39$

(b) $K = 6, |P_j| = 4$
$P = 21, |\mathcal{X}| = 63$

(c) $K = 4, |P_j| = 8$
$P = 25, |\mathcal{X}| = 125$

Figure 5: Excerpts of frames from our synthetic video sequences, and the features of the corresponding articulated objects (number of arms $|P_j|$, $j > 1$, length of arms $K - 1$, total number of parts $P$, and dimension of state space $\mathcal{X}$).

Table 1: Resampling times (in seconds) for the estimation of the density of different objects, with $N = \{100, 600\}$.

|  | Fig. 5.a | | Fig. 5.b | | Fig. 5.c | |
|---|---|---|---|---|---|---|
|  | 100 | 600 | 100 | 600 | 100 | 600 |
| Multinomial | 0.5 | 17.1 | 1.3 | 46.9 | 1.8 | 79.6 |
| Systematic | 0.5 | 19.8 | 1.2 | 53.6 | 1.7 | 80.5 |
| Stratified | 0.5 | 16.9 | 1.3 | 44.8 | 1.7 | 74.9 |
| Residual | 0.5 | 20.3 | 1.3 | 55.7 | 1.8 | 83.4 |
| Weighted | 1.0 | 33.0 | 2.5 | 90.1 | 3.5 | 157.8 |
| Combinatorial | 0.7 | 10.6 | 1.5 | 26.3 | 1.5 | 22.3 |

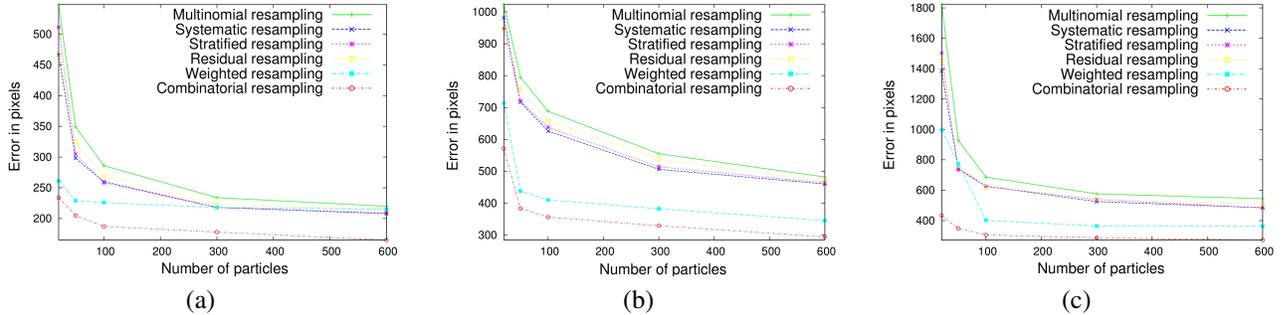

Figure 6: Comparison of convergence for different resampling approaches: errors of estimation of the density of objects depending on $N$: (a) with $|P_i| = 4$, $K = 4$ (object of Fig. 5.(a)), (b) with $|P_i| = 4$, $K = 6$ (object of Fig. 5.(b)) and (c) with $|P_i| = 8$, $K = 4$ (object of Fig. 5.(c)).

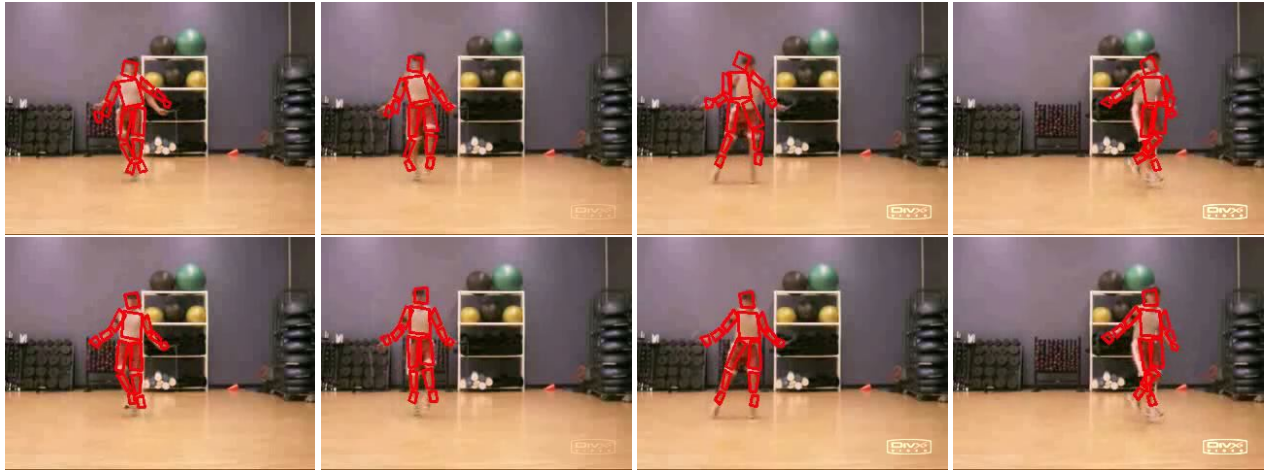

Figure 7: Tracking results on `JumpRope` sequence with $N = 500$ particles (frames 10, 50, 121 and 234). First line, using residual resampling, last line, using our combinatorial resampling.

of the densities to estimate. This feature holds even when there are wide movements over time and when images have a low resolution. Qualitative results are given by superimposing on the frames of the sequences a red articulated object corresponding to the estimation derived from the weighted sum of the particles. For this test, we fixed $\sigma_x = \sigma_y = 2$ pixels and $\sigma_\theta = 0.08$ rad.

Figure 7 shows tracking results on the `JumpRope` sequence (containing 290 frames of $320 \times 240$ pixels) with $N = 500$ particles. In this sequence, a person is quickly moving from left to right while jumping, and crossing/uncrossing his arms and legs. For this test, we defined an articulated object with $P = 12$ parts, hence $|\mathcal{X}| = 36$, and we compared the estimations resulting from PS with a residual resampling (top line) with those resulting from our proposed resampling approach (bottom line). As can be observed, our approach produces better results: its estimations are more stable along the sequence. For example, on the images of the 2nd and 3rd columns, we can see the estimation of the articulated object fails with residual resampling but is correct with our combinatorial resampling. For this sequence, on average over 20 runs, our method needed 16 seconds while residual resampling needed 22. In addition, our algorithm proved to be more robust and provided more accurate results. As for synthetic sequences, our tests show that the higher the number of particles, the more our algorithm outperforms residual resampling in terms of response time. It is also always more accurate.

## 6 CONCLUSION

In this paper, we have introduced a new resampling method called *Combinatorial Resampling*. From a given sample $S$, this algorithm constructs implicitly a new sample $S'$ exponentially larger than $S$. By construction, $S'$ is more representative than $S$ of the density over the whole state space and resampling from $S'$ rather than $S$ produces much better results, as confirmed by our experiments. We proved the mathematical correctness of the method and showed that it is effective for real time tracking. For future researches, there remains to exhibit theoretical convergence results for PS combined with this new resampling scheme.